\documentclass[11pt,a4paper]{article}
\usepackage[hyperref]{acl2019}
\usepackage{times}
\usepackage{latexsym}
\usepackage{graphicx}
\usepackage{wrapfig}
\usepackage{booktabs}

\usepackage{url}

\aclfinalcopy 

\definecolor{cb_purple}{rgb}{0.6, 0.31, 0.64}

\title{GLTR:  Statistical Detection and Visualization of Generated Text}

\author{Sebastian Gehrmann \\
  Harvard SEAS\\
  \small{\texttt{gehrmann@seas.harvard.edu}} \\\And
  Hendrik Strobelt \\
  IBM Research\\
  MIT-IBM Watson AI lab\\
  \small{\texttt{hendrik.strobelt@ibm.com}} \\\And
  Alexander M. Rush \\
  Harvard SEAS \\
  \small{\texttt{srush@seas.harvard.edu}}
  }

\date{}

\begin{document}
\maketitle

\begin{abstract}
The rapid improvement of language models has raised the specter of abuse of text generation systems. This progress motivates the development of simple methods for detecting generated text that can be used by and explained to non-experts. We develop GLTR, a tool to support humans in detecting whether a text was generated by a model. GLTR applies a suite of baseline statistical methods that can detect generation artifacts across common sampling schemes. In a human-subjects study, we show that the annotation scheme provided by GLTR improves the human detection-rate of fake text from 54\% to 72\% without any prior training. GLTR is open-source and publicly deployed, and has already been widely used to detect generated outputs.\looseness=-1
\end{abstract}

\section{Introduction}

The success of pretrained language models for natural language understanding~\cite{mccann2017learned,devlin2018bert,peters2018deep} has led to a race to train unprecedentedly large language models~\cite{radford2019language}. These large language models have the potential to generate textual output that is indistinguishable from human-written text to a non-expert reader. That means that the advances in the development of large language models also lower the barrier for abuse.


Instances of malicious autonomously generated text at scale are rare but often high-profile, for instance when a simple generation system was used to create fake comments in opposition to net neutrality~\citep{grimaldi_2018}.
Other scenarios include the possibility of generating false articles~\citep{wang2017liar} or misleading reviews~\citep{fornaciari2014identifying}. Forensic techniques will be necessary to detect this automatically generated text.
These techniques should be accurate, but also easy to convey to non-experts and require little setup cost.

\begin{figure}[t]
    \centering
    \includegraphics[width=0.48\textwidth]{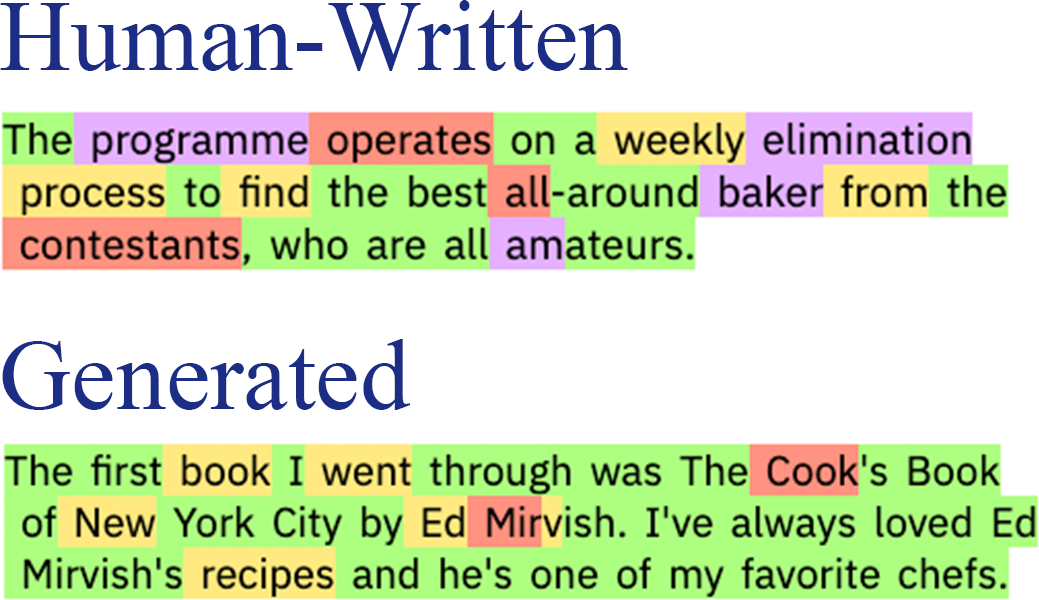}
    \caption{The top-k overlay within GLTR. It is easy to distinguish sampled from written text. The real text is from the Wikipedia page of The Great British Bake Off, the fake from GPT-2 large with temperature 0.7.}
    \label{fig:overlay}
\end{figure}

\begin{figure*}[t!]
    \centering
    \includegraphics{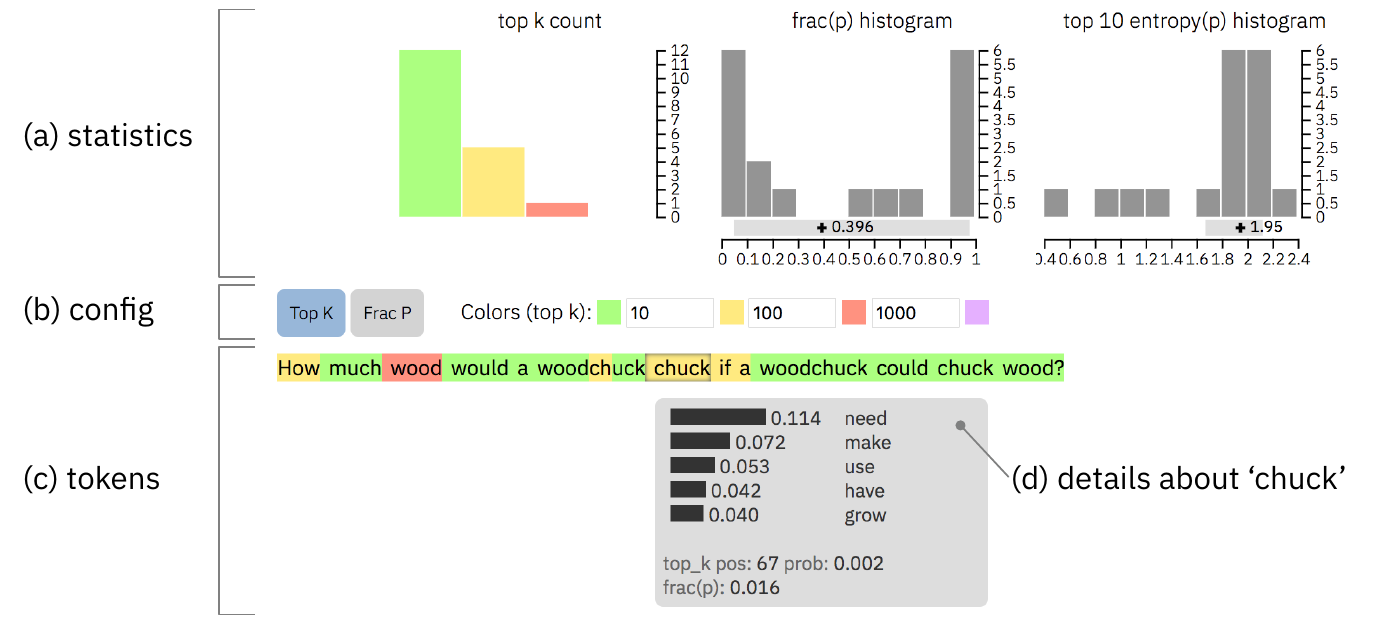}
    \caption{User interface for GLTR. On the top, we show three graphs with global information (a). Below the graphs, users can switch between two different annotations and customize the top-k thresholds (b). On the bottom, each token is shown with the associated annotation as heatmap (c). The tooltip (d) highlights information about the current prediction when hovering over the word ``chuck''.}
    \label{fig:teaser}
\end{figure*}

In this work, we argue that simple statistical detection methods for generated/fake text can be applied within a visual tool to assist in detection. The underlying assumption is that systems over generate from a limited subset of the true distribution of natural language, for which they have high confidence. In a white-box setting where we have access to the system distribution, this property can be detected by computing the model density of generated output and comparing it to human-generated text. We further hypothesize that these methods generalize to black-box scenarios, as long as the fake text follows a similar sampling assumption and is generated by a large language model.


We develop a visual tool, GLTR, that highlights text passages based on these metrics, as shown in Figure~\ref{fig:overlay}\footnote{Our tool is available at \url{http://gltr.io}. \\The code is provided at \url{https://github.com/HendrikStrobelt/detecting-fake-text}}.
We conduct experiments to empirically test these metrics on a set of widely-used language models and show that real text uses a wider subset of the distribution under a model. This is noticeable especially when the model distribution is low-entropy and concentrates most probability in a few words. We demonstrate in a human-subjects study that without the tool, subjects can differentiate between human- and model-generated text only 54\% of the time. With our tool, subjects were able to detect fake text with an accuracy of over 72\% without any prior training. By presenting this information visually, we also hope the tool teaches users to notice the artefacts of text generation systems.\looseness=-1


\section{Method}
\label{sec:analy}

Consider the generation detection task as deciding whether a sequence of words $\hat{X}_{1:N}$ have been written by a human or generated from a model. We do not have supervision for this task, and instead, want to use distributional properties of the underlying language. In the white-box case, we are also given full access to the language model distribution, $p(X_i \ | X_{1:i-1})$, that was used in generation. In the general case, we assume access to a different learned model of the same form. This approach can be contextualized in the evaluation framework proposed by \citet{hashimoto2019unifying} who find that human-written and generated text can be discriminated based on the model likelihood if the human acceptability is high. 

The underlying assumption of our methods is that to generate natural looking text, most systems sample from the head of the distribution, e.g., through max sampling~\citep{gu2017trainable}, k-max sampling~\citep{fan2018hierarchical}, beam search~\citep{chorowski2016towards,shao2017generating}, temperature-modulated sampling~\citep{dagan1995selective}, or even implicitly with rule-based templated approaches. These techniques are biased, but seem to be necessary for fluent output and are widely used. We therefore propose three simple tests, using a detection model, to assess whether text is generated in this way:  \textbf{(Test~1)} the probability of the  word, e.g. $p_{\mathrm{det}}(X_i=\hat{X}_i | X_{1: i-1})$,   \textbf{(Test~2)} the absolute rank of a word, e.g. rank in $p_{\mathrm{det}}(X_i |X_{1:i-1})$, and \textbf{(Test~3)} the entropy of the predicted distribution, e.g.  $-\sum_w p_{\mathrm{det}}(X_i = w |X_{1:i-1} ) \log p_{\mathrm{det}}(X_i=w |X_{1:i-1} )$. The first two test whether a generated word is sampled from the top of the distribution and the last tests whether the previously generated context is well-known to the detection system such that it is (overly) sure of its next prediction.

\begin{figure*}[t]
    \centering
    \includegraphics{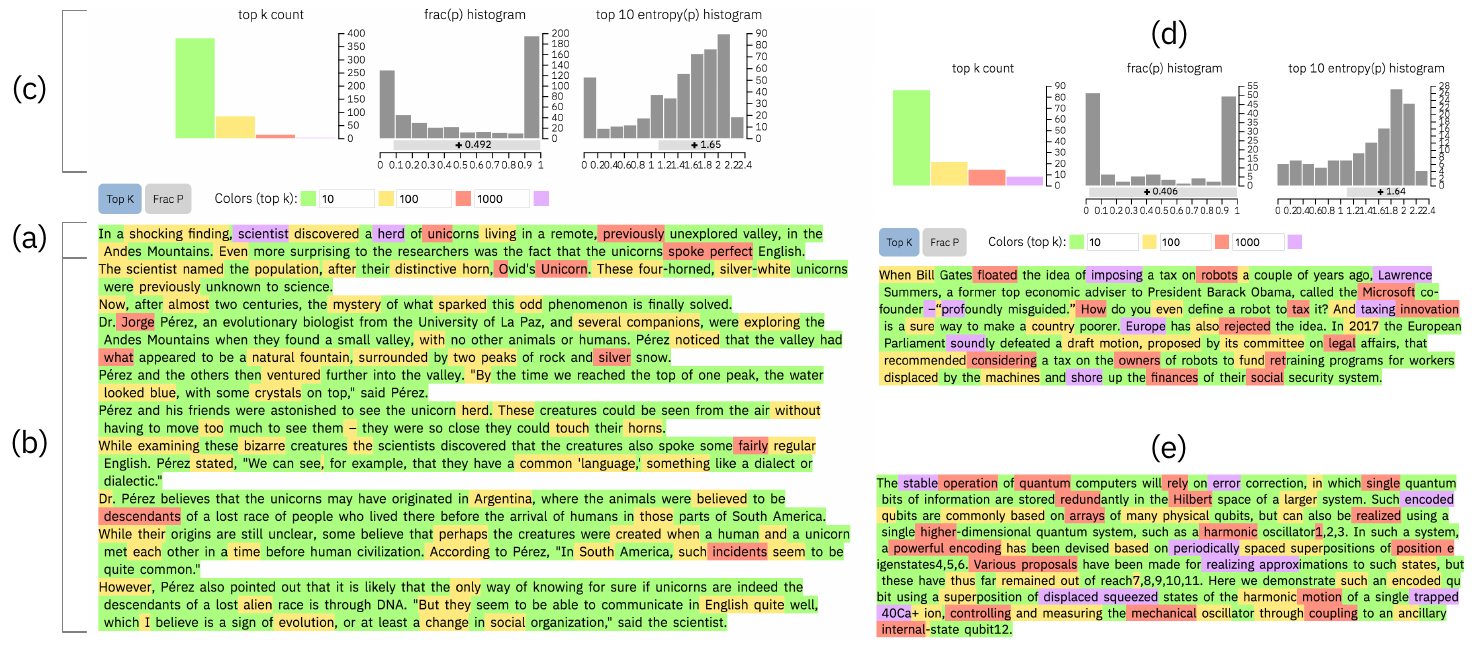}
    \caption{On the left, we analyze a generated sample (a-c) with GLTR that is generated from a non-public GPT-2 model. The first sentence (a) is the prompt given to the model. We can observe that the generated text (b) is mostly highlighted in green and yellow, which strongly hints at a generated text. The histograms (c) show additional hints at the automatic generation. 
    On the right, we show samples from a real NYT article (d) and a scientific abstract (e). Compared to the "unicorn" example, the fraction of red and purple words is much higher.
    }
    \label{fig:examples}
\end{figure*}

\section{GLTR: Visualizing Outliers}
We apply these tests within our tool GLTR (pronounced Glitter) -- a \textbf{G}iant \textbf{L}anguage model \textbf{T}est \textbf{R}oom. GLTR aims to both teach users what to be aware of when assessing whether a text is real, and to assist them in performing forensic analyses. It works on a per-instance basis for \textit{any} textual input.

The backend supports multiple detection models. Our publicly deployed version uses both BERT~\citep{devlin2018bert} and GPT-2 117M~\citep{radford2019language}. 
Since GPT-2 117M is a standard left-to-right language model, we compute $p_{\mathrm{det}}(X_i\ |\ X_{1\ldots i-1})$ at each position $i$ in a text $X$.
BERT is trained to predict a masked token, given a bidirectional context. Thus, we iteratively mask out each correct token $\hat{X}_i$ and use a context of $30$ words to each side as input to estimate $p_{\mathrm{det}}(X_i|X_{i-30\ldots i-1}, X_{i+1\ldots X_i+30})$\footnote{While BERT can handle inputs of length 512, we observed only minor differences between using the full and shortened contexts.}.


The central feature of the tool is the overlay function, shown in Figure~\ref{fig:teaser}c, which can render arbitrarily chosen top-k buckets (Test-2) as an annotation over the text.
By default, a word that ranks within the top 10 is highlighted in green, top 100 in yellow, top 1,000 in red, and the rest in purple. GLTR also supports an overlay for Test-1 that highlights the probability of the chosen word in relation to the one that was assigned the highest probability. Since the two overlays provide evidence from two separate sources, their combination helps to form an informed assessment.

The top of the interface (Figure~\ref{fig:teaser}a), shows one graph for each of the three tests. The first one shows the distribution over the top-k buckets, the second the distribution over the values from the second overlay, and the third the distribution over the entropy values.
For a more detailed analysis, hovering over a word (Figure~\ref{fig:teaser}d) shows a tooltip with the top 5 predictions, their probabilities, and the rank and probability of the following word.

The backend of GLTR is implemented in PyTorch and is designed to ensure extensibility. New detection models can be added by registering themselves with the API and providing a model and a tokenizer. This setup will allow the front-end of the  tool to continue to be used as improved language models are released.


\paragraph{Case Study}

We demonstrate the functionality of GLTR by analyzing three samples from different sources, shown in Figure~\ref{fig:examples}. The interface shows the results of detection analysis with GPT-2 117M. The first example is generated from GPT-2 1.5B.
Here the example is conditioned on a seed text.\footnote{\textit{In a shocking finding, scientist discovered a herd of unicorns living in a remote, previously unexplored valley, in the Andes Mountains. Even more surprising to the researchers was the fact that the unicorns spoke perfect English}} The analysis shows that not a single token in the generated text is highlighted in purple and very few in red. Most words are green or yellow, indicating high rank. Additionally, the second histogram shows a high fraction of high-probability choices. A final indicator is the regularity in the third histogram with a high fraction of low-entropy predictions and an almost linear increase in the frequency of high-entropy words.

In contrast, we show two human-written samples; one from a New York Times article and a scientific abstract (Figure~\ref{fig:examples}d+e). There is a significantly higher fraction of red and purple (e.g. non-obvious) predictions compared to the generated example. The difference is also observable in the histograms where the fraction of low-probability words is higher and low-entropy contexts smaller.

\section{Empirical Validation}

We validate the detection features by comparing 50 articles for each of 3 generated and 3 human data sources. The first two sources are documents sampled from \emph{GPT-2 1.5B} \cite{radford2019language}. We use a random subset of their released examples that were generated (1) with a temperature of 0.7 and (2) truncated to the top 40 predictions. As alternative source of generated text, we take articles that were generated by the autonomous Washington Post \emph{Heliograf} system, which covers local sports results and gubernatorial races.
As human-written sources, we choose random paragraphs from the bAbI task children book corpus (CBT)~\citep{hill2015goldilocks}, New York Times articles (NYT), and scientific abstracts from the journals nature and science (SA). To minimize overlap with the training set, we constrained the samples to publication dates past or close to the release of the GPT-2 models.

Our first model uses the average probability of each word in a document as single feature (Test 1) and the second one the distribution over four buckets (highlight colors in GLTR) of absolute ranks of predictions (Test 2). As a baseline we consider a logistic regression over a bag-of-words representation of each document. We cross-validate the results by training on each combination of four of the sources (two real/fake) and testing on the remaining two.

\paragraph{Results}

\begin{table}[t!]
\begin{center}
\begin{tabular}{@{}lc@{}}
\toprule
Feature & AUC \\ \midrule
Bag of Words & 0.63 $\pm$0.11 \\ \midrule
(Test 1 - GPT-2) Average Probability & 0.71 $\pm$0.25 \\
(Test 2 - GPT-2) Top-K Buckets & 0.87 $\pm$0.07 \\ \midrule
(Test 1 - BERT) Average Probability & 0.70 $\pm$0.27 \\
(Test 2 - BERT) Top-K Buckets & 0.85 $\pm$0.09 \\ \bottomrule
\end{tabular}
\end{center}
\caption{Cross-validated results of fake-text discriminators. Distributional information yield a higher informativeness than word-features in a logistic regression.}
\label{tab:lr}
\end{table}

\begin{figure}[t]
    \centering
    \includegraphics[width=1.05\linewidth]{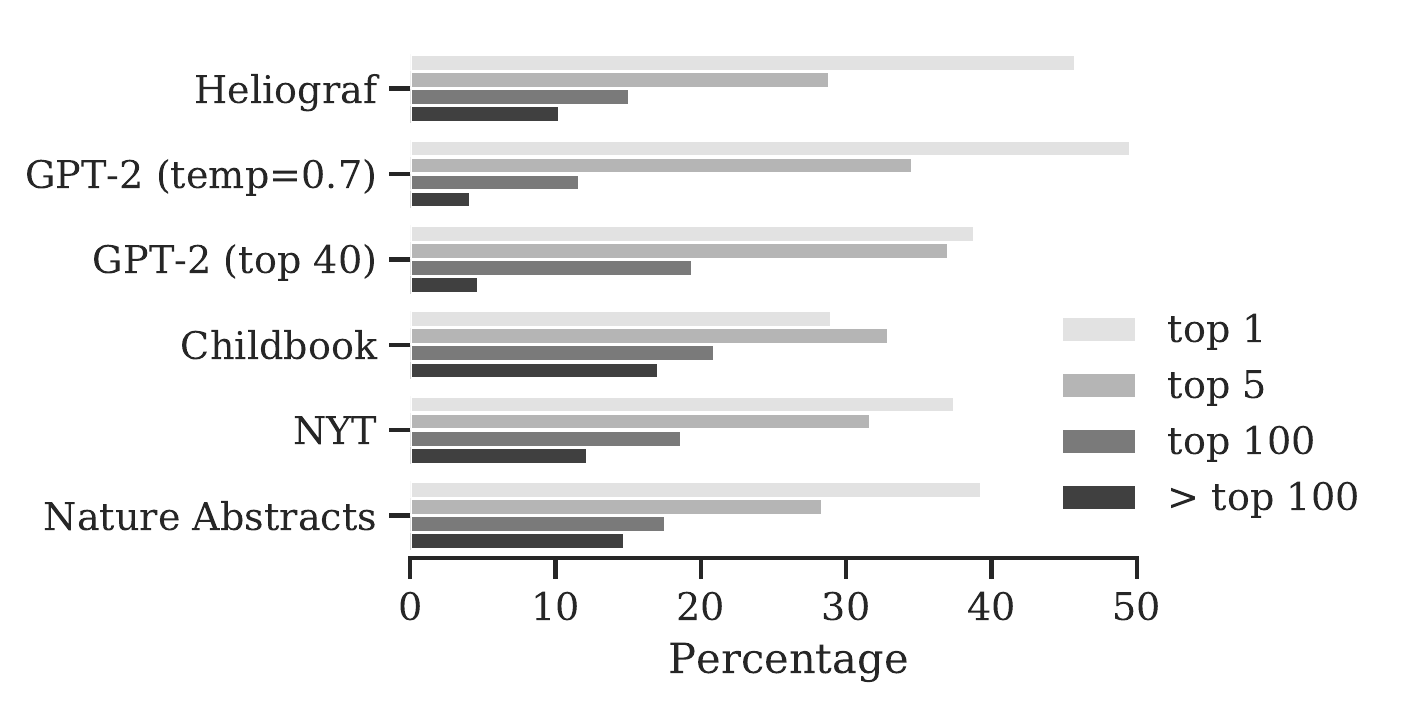}
    \caption{Distribution over the rankings of words in the predicted distributions from GPT-2. The real text in the bottom three examples has a consistently higher fraction of words from the tail of the distribution.
    }
    \label{fig:topk}
    \includegraphics[width=0.49\textwidth]{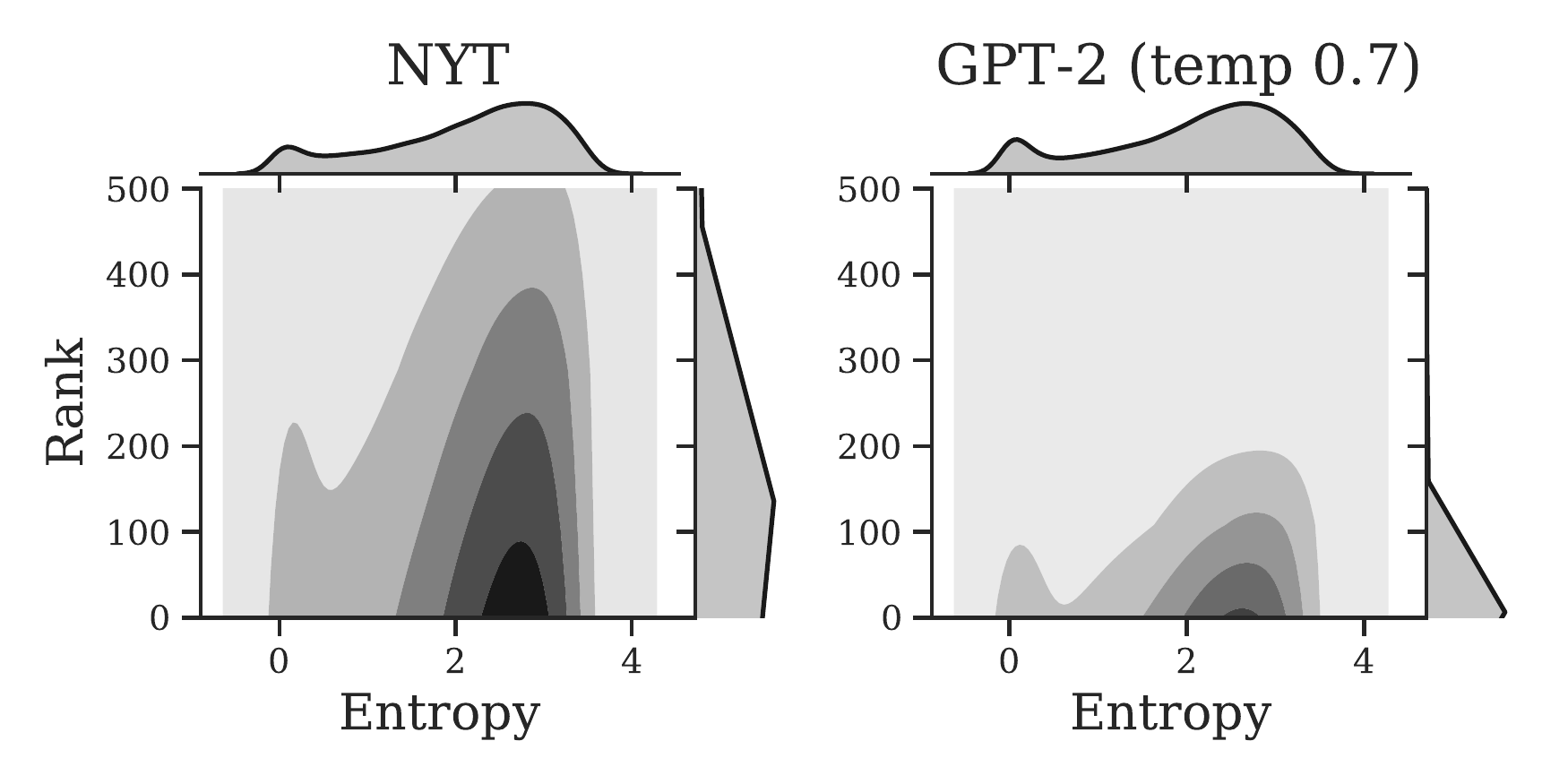}
    \caption{A kernel density estimate of the contextual entropy (Test 3) versus the next-word rank (Test 2) for NYT and GPT-2. Human-written text (NYT) is more likely to have high-rank words, even in low-entropy contexts.}
    \label{fig:kde}
\end{figure}

As Table~\ref{tab:lr} illustrates, the GLTR features lead to better separation than word-features, both with and without access to the true generating model.
The classifier that uses ranking information learns that real text samples from the tail of the distribution more frequently. The odds ratio for a word outside the top 100 predictions is $5.32$, while the odds ratio for being the top 1 prediction is $0.09$.
Figure~\ref{fig:topk} presents the distribution of rankings under GPT-2 and further corroborates this finding. Real texts use words outside of the top 100 predictions $2.41$ times as frequently under GPT-2 ($1.67$ for BERT) as generated text, even compared to sampling with a lower temperature.

To get a better sense of how low-rank words enter into natural text,
we look at the probability of each word compared to its relative rank.
We hypothesize that human authors use low-rank words, even when the entropy is low, a property that sampling methods for generated text avoid.
We compare the relationship of the entropy and rank of the next word by computing a Gaussian Kernel-density estimate over their distributions. As shown in Figure~\ref{fig:kde}, human text uses high-rank words more frequently, regardless of the estimated entropy.

\section{Human-Subjects Study}

To evaluate the efficacy of the GLTR tool, we conducted a human-subjects study on 35 volunteer students in a college-level NLP class. Our goal was to both have
students be able to tell generated text from real, but also to see which parts raised the suspicion of the students. In two rounds, students were first shown five texts without overlay and then five texts with overlay and were asked to assess which texts were real within 90 seconds. In between the rounds, we presented a brief tutorial on the overlay and showed the example in Figure~\ref{fig:overlay} but did not disclose any information about the study. For each participant and round, we presented two texts generated from GPT-2 with 0.7 temperature, one from Heliograf, and two from NYT.\footnote{We randomly sampled one paragraph of text and resampled NYT if it was covering recent, well-known events.}
We alleviated bias from the text selection by randomly assigning texts to either of the two rounds between students.

\paragraph{Results}

The results demonstrate the ease of use of the overlay. Without the interface, the participants achieved an accuracy of 54.2\%, barely above random chance. While only 40\% of texts were real, they trusted 56.0\% of texts, Heliograf at a higher rate than GPT-2 (68.6\% vs. 51.4\%, $p<0.01$). The difficulty of the task without overlay was rated at 3.89 on a 5-point Likert scale, further supporting the need for assistive systems.
With the interface, the performance improved to 72.3\%. The average treatment effect shows an improvement of 18.1\% with $p<0.001$, even after controlling for whether a participant is a native speaker and how difficult they rated the task. $42.1$\% of the participants stated that the interface helped them be more accurate, and $37.1$\% found that it helped them to identify fakes faster.

\paragraph{Qualitative Findings}
The tool caused students to think about the properties of the fake text. While humans would vary expressions in real texts, models rarely generate synonyms or referring expressions for entities, which does not follow the theory of centering in discourse analysis~\citep{grosz1995centering}.
An example of this is shown in the text in Figure~\ref{fig:examples}b in which the model keeps generating the name P\'erez and never refers to him as “he”.
Another observation was that samples from Heliograf exhibit high parallelism in sentence structure. Since previous work has found that neural language models learn long linguistic structures as well, we imagine that sentence structure analysis can further be used for forensic analysis. We hope that automatic analysis and visualization like GLTR will help students better understand the generation artifacts in current systems.

\section{Related Work}

While statistical detection methods have been applied in the past, the increase in language model power upends past assumptions in this area. \citet{lavergne2008detecting} introduce prediction entropy as an indicator of fake text. However, their findings are the opposite of ours (low entropy for generated text), a change which is indicative of language model improvements. Similar work finds that texts differ in perplexity under a language model~\citep{beresneva2016computer}, frequency of rare bigrams~\citep{grechnikov2009detection}, and n-gram frequencies~\citep{badaskar2008identifying}. Similar methods that detect machine translation~\citep{arase2013machine}. \citet{hovy2016enemy} finds that a logistic regression model can detect generated product reviews at a higher rate than human judges, indicating that humans struggle with this task. 
Finally, we distinguish this task from detecting misinformation in text~\citep[e.g.][]{shu2017fake}. We aim to understand the statistical signature and not the content of text.

\section{Discussion and Conclusion}

We show how detection models can be applied to analyze whether a text is automatically generated using only simple statistical properties. We apply the insights from the analysis to build GLTR, a tool that assists human readers and improves their ability to detect fake texts.

\paragraph{Impact} GLTR aims to educate and raise awareness about generated text. To explain GLTR to non-NLP experts, we included a blog post on the web page with examples and an explanation of GLTR. Within the first month, GLTR had 30,000 page views for the demo and 21,000 for the blog. Numerous news websites and policy researchers reached out to discuss the ethical implications of language generation. The feedback from these discussions and in-person presentations helped us to refine our publicly released examples and explore the limits of our detection methods.

\paragraph{Future Work}
A core assumption of GLTR is that systems use biased sampling for generating text. One can imagine adversarial schemes that aim to fool our overlay; however, forcibly sampling from the tail decreases the coherence of a text which may make it harder to fool human readers. Another potential limitation are samples conditioned on a hidden seed text. A conditional distribution will look different, even if we have access to the model. Our preliminary qualitative investigations with GLTR show a relatively short-range memory on this seed, but it is crucial to conduct more in-depth evaluations on the influence of conditions in future work. The findings further motivate future work on how to use our methods as part of autonomous classifiers to assist moderators on social media or review platforms.


\section*{Acknowledgments}

AMR gratefully acknowledges the support of NSF 1845664 and a Google research award. 

\bibliography{acl2019}
\bibliographystyle{acl_natbib}

\end{document}